\documentclass[runningheads]{llncs}
\usepackage[T1]{fontenc}
\usepackage{graphicx,verbatim}
\usepackage{mathtools}
\usepackage{hyperref}
\usepackage{booktabs}
\usepackage{multicol}
\usepackage{multirow}
\usepackage{amssymb}
\usepackage{pifont}
\newcommand{\xmark}{\ding{55}}%
\usepackage[table]{xcolor}
\definecolor{lightgray}{gray}{0.93}
\begin{document}
\title{An Artifact-based Agent Framework for Adaptive and Reproducible Medical Image Processing}
\titlerunning{Adaptive and Reproducible Medical Image Processing}
%
\author{
Lianrui~Zuo$^{1}$\orcidID{0000-0002-5923-9097},
Yihao~Liu$^{1}$\orcidID{0000-0003-3187-9903},
Gaurav~Rudravaram$^{1}$\orcidID{0009-0001-9087-462X},
Karthik~Ramadass$^{1,2}$\orcidID{0000-0002-4610-3860},
Aravind~R.~Krishnan$^{1}$\orcidID{0009-0000-1829-8245},
Michael~D.~Phillips$^{1}$,
Yelena~G.~Bodien$^{4}$,
Mayur~B.~Patel$^{4}$,
Paula~Trujillo$^{5}$,
Yency~Forero~Martinez$^{6}$,
Stephen~A.~Deppen$^{7,11}$,
Eric~L.~Grogan$^{7}$,
Fabien~Maldonado$^{7}$,
Kevin~McGann$^{7}$,
Hudson~M.~Holmes$^{8}$,
Laurie~E.~Cutting$^{12,13}$,
Yuankai~Huo$^{2}$\orcidID{0000-0002-2096-8065},
Bennett~A.~Landman$^{1,2,3,9,10}$\orcidID{0000-0001-5733-2127}
}

\authorrunning{Zuo~et~al.}

\institute{
$^{1}$Dept. of Electrical and Computer Engineering, Vanderbilt University, Nashville, TN, USA\\
$^{2}$Dept. of Computer Science, Vanderbilt University, Nashville, TN, USA\\
$^{3}$Dept. of Biomedical Engineering, Vanderbilt University, Nashville, TN, USA\\
$^{4}$Dept. of Surgery, Vanderbilt University Medical Center, Nashville, TN, USA\\
$^{5}$Dept. of Neurology, Vanderbilt University Medical Center, Nashville, TN, USA\\
$^{6}$Dept. of Medicine, Vanderbilt University Medical Center, Nashville, TN, USA\\
$^{7}$Dept. of Thoracic Surgery, Vanderbilt University Medical Center, \\Nashville, TN, USA\\
$^{8}$Dept. of Biomedical Informatics, Vanderbilt University Medical Center, \\ Nashville, TN, USA\\
$^{9}$Dept. of Radiology and Radiological Sciences, Vanderbilt University Medical Center, Nashville, TN, USA\\
$^{10}$Vanderbilt University Institute of Imaging Science, Nashville, TN, USA\\
$^{11}$Veteran Affairs, Tennessee Valley Healthcare System, Nashville, TN, USA\\
$^{12}$Peabody College, Vanderbilt University, Nashville, TN, USA\\
$^{13}$Vanderbilt Brain Institute, Vanderbilt University, Nashville, TN, USA \\
\email{lianrui.zuo@vanderbilt.edu}}

\maketitle              
\begin{abstract}
Medical imaging research is increasingly shifting from controlled benchmark evaluation toward real-world clinical deployment.
In such settings, applying analytical methods extends beyond model design to require dataset-aware workflow configuration and provenance tracking.
Two requirements therefore become central: \textbf{adaptability}, the ability to configure workflows according to dataset-specific conditions and evolving analytical goals; and \textbf{reproducibility}, the guarantee that all transformations and decisions are explicitly recorded and re-executable.
Here, we present an artifact-based agent framework that introduces a semantic layer to augment medical image processing. 
The framework formalizes intermediate and final outputs through an artifact contract, enabling structured interrogation of workflow state and goal-conditioned assembly of configurations from a modular rule library. 
Execution is delegated to a workflow executor to preserve deterministic computational graph construction and provenance tracking, while the agent operates locally to comply with most privacy constraints.
We evaluate the framework on real-world clinical CT and MRI cohorts, demonstrating adaptive configuration synthesis, deterministic reproducibility across repeated executions, and artifact-grounded semantic querying. 
These results show that adaptive workflow configuration can be achieved without compromising reproducibility in heterogeneous clinical environments.
\keywords{CT  \and MRI \and Clinical translation \and Agent}

\end{abstract}
\section{Introduction}
Methodological progress in medical imaging has been driven by advances in model architectures and learning paradigms, typically evaluated on curated benchmark datasets or structured research cohorts~\cite{ferrucci2023baltimore,lamontagne2019oasis,naseer2022performance}. Public initiatives and grand challenges have further accelerated innovation by providing standardized datasets and controlled evaluation protocols~\cite{chen2025beyond,menze2014multimodal,zhang2024harmonization}. These settings enable rigorous comparison of methods under well-defined experimental conditions.

\begin{figure}[!tb]
    \centering
    \includegraphics[width=0.99\linewidth]{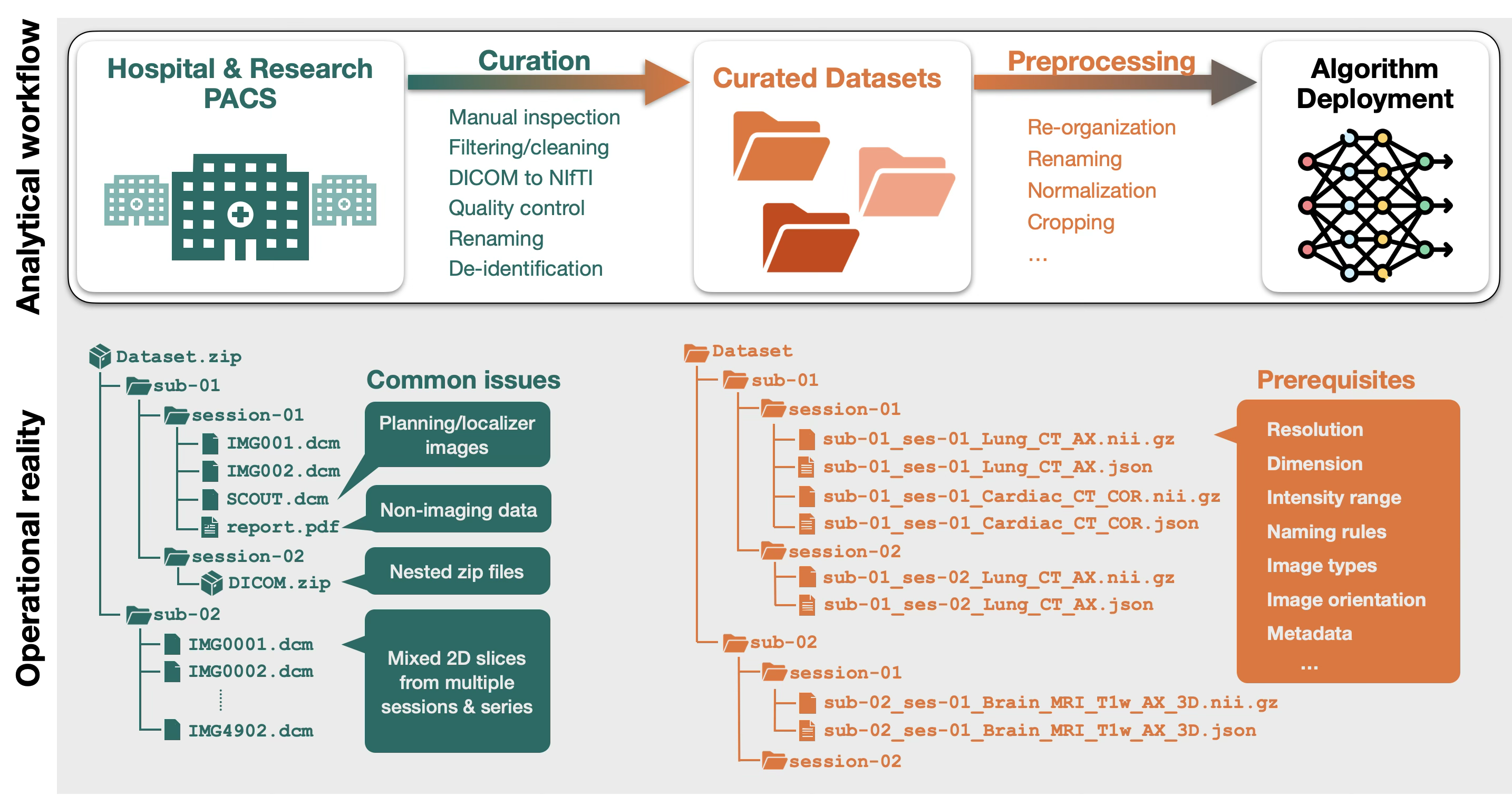}
    \caption{The typical analytical workflow and operational reality in imaging research. Imaging data originate in hospital and research PACS, where heterogeneous file structures, mixed modalities, and nested archives are common. Extensive curation and processing are often required before most analytical algorithms can be deployed.}
    \label{fig:messy_datasets}
\end{figure}

In contrast, when the goal becomes applying existing analytical methods to a new research or clinical dataset, the operational landscape changes substantially.
Imaging data originate in hospital and research picture archiving and communication systems~(PACS), where heterogeneous file organization, mixed modalities, nested archives, and acquisition planning images are common~(Figure~\ref{fig:messy_datasets}). 
For example, a single subject folder very often contains raw DICOM slices mixed with localizer images, reformatted series, and non-imaging documents, without standardized naming or directory structure.
Transforming such data into analyzable form requires multi-stage curation, format conversion, quality control, and preprocessing. 
These steps often demand considerable manual and cohort-specific effort.
As a result, workflow assembly becomes an expertise-driven process that is difficult to generalize and reproduce.

These challenges are further compounded in clinical environments governed by privacy regulations and data use agreements. 
In many settings, raw images or metadata cannot be transmitted to external settings. 
Ironically, the scenarios that require the greatest interactive effort for dataset adaptation are precisely those in which cloud-based models may be constrained or require additional regulatory and institutional approvals. 
Intelligent support systems must therefore be designed to operate within such constraints while preserving deterministic execution and traceable provenance.

To this end, two key requirements emerge. First is \underline{adaptability}: the ability to reason about heterogeneous datasets and evolving analytical goals to configure appropriate processing workflows. 
Second is \underline{reproducibility}: the guarantee that all transformations and decisions are explicitly represented and deterministically re-executable~\cite{peng2011reproducible,stodden2016enhancing}. 
Existing workflow engines~\cite{koster2012snakemake,di2017nextflow} ensure deterministic execution once a workflow is defined, and standardization initiatives~\cite{poldrack2024past} improve structural consistency. 
However, these tools assume pre-specified configurations and do not provide mechanisms for dataset-aware adaptation or semantic inspection of intermediate workflow state.

To bridge this gap, we propose an artifact contract-based agent framework that introduces a semantic layer above deterministic workflow execution. 
The framework defines an artifact contract that encodes all workflow outputs and states~(intermediate and final) into structured, queryable records with explicit provenance. 
A semantic query module operates strictly over these contract-compliant artifacts to support dataset-level inspection.
For analytical objectives, a workflow assembly module synthesizes goal-conditioned configurations from a modular rule library, enabling dataset-aware adaptation while preserving deterministic execution through an underlying workflow executor. 
The entire system is deployed fully locally to comply with regulations in many clinical environments.
The contributions are threefold:
\begin{itemize}
    \item An artifact-based contract that formalizes workflow state for structured and auditable reasoning;
    \item A constrained agent layer enabling goal-conditioned workflow assembly and artifact-grounded semantic querying;
    \item Empirical validation across clinical CT and MRI cohorts demonstrating adaptive configuration synthesis with deterministic reproducibility.
\end{itemize}

\section{Methods}
\subsection{Problem Setup}
\subsubsection{Dataset, Workflow, and Artifact Contract}
Let $\mathcal{D} = \{d_i \}_{i=1}^N$ denote a dataset consisting of $N$ subjects, where each $d_i$ may include multiple longitudinal imaging sessions, file types, and associated metadata. 
Let $\mathcal{S} = \{ s_k \}_{k=1}^{K}$ denote a library of modular processing rules that a user can choose. 
Each rule $s_k$ specifies a processing step and its dependencies, such as file format conversion or a segmentation algorithm.
A workflow configuration is defined as $C = (\pi, \theta)$, where $\pi \subseteq \mathcal{S}$ represents an ordered subset of selected rules together with their dependency graph, and $\theta$ denotes rule-level parameters. 
Given dataset $\mathcal{D}$ and configuration $C$, execution is performed by a deterministic workflow executor
$P (\mathcal{D}, C) = \mathcal{A}$,
where $\mathcal{A} = \{ a_j \}_{j=1}^{M}$
denotes the set of artifacts produced during execution. 
Artifacts include intermediate and final outputs of the workflow, including transformed images, derived measurements (e.g., classification labels), quality control indicators, and log records.
Conventional workflow engines guarantee reproducibility under fixed configuration $C$, but assume that configurations are manually specified and that interpretation of workflow state occurs externally. 
They do not provide structured mechanisms for adaptive configuration to new datasets or semantic interrogation of intermediate state.

\noindent \textbf{Agentic Systems}
Recent advances in large language models (LLMs) have enabled agentic systems that perform goal-directed reasoning through iterative planning and tool invocation~\cite{yao2022react,schick2023toolformer,park2023generative}. 
In a general agent architecture, a planner selects actions from a tool set conditioned on current state and user objectives, executes those actions, and updates memory based on resulting observations~\cite{schick2023toolformer}.
In the context of medical image processing, however, unrestricted tool invocation is insufficient: execution must remain deterministic and reproducible.

Within our formulation, the dataset $\mathcal{D}$ and artifact set $\mathcal{A}$ define the observable state; the rule library $\mathcal{S}$ defines the available tools~\cite{schick2023toolformer}; and the configuration $C = (\pi, \theta)$ represents a structured plan over $\mathcal{S}$. 
We therefore seek a constrained agentic system where planning operates over defined workflow components, memory is grounded in contract-compliant artifacts $\mathcal{A}$~\cite{park2023generative}, and execution is delegated to the workflow executor $P(\mathcal{D},C)$. 
This enables adaptability through configuration synthesis while preserving reproducibility through deterministic execution.

\subsection{Proposed Framework}
\noindent \textbf{Overview}
\begin{figure}[!tb]
    \centering
    \includegraphics[width=0.99\linewidth]{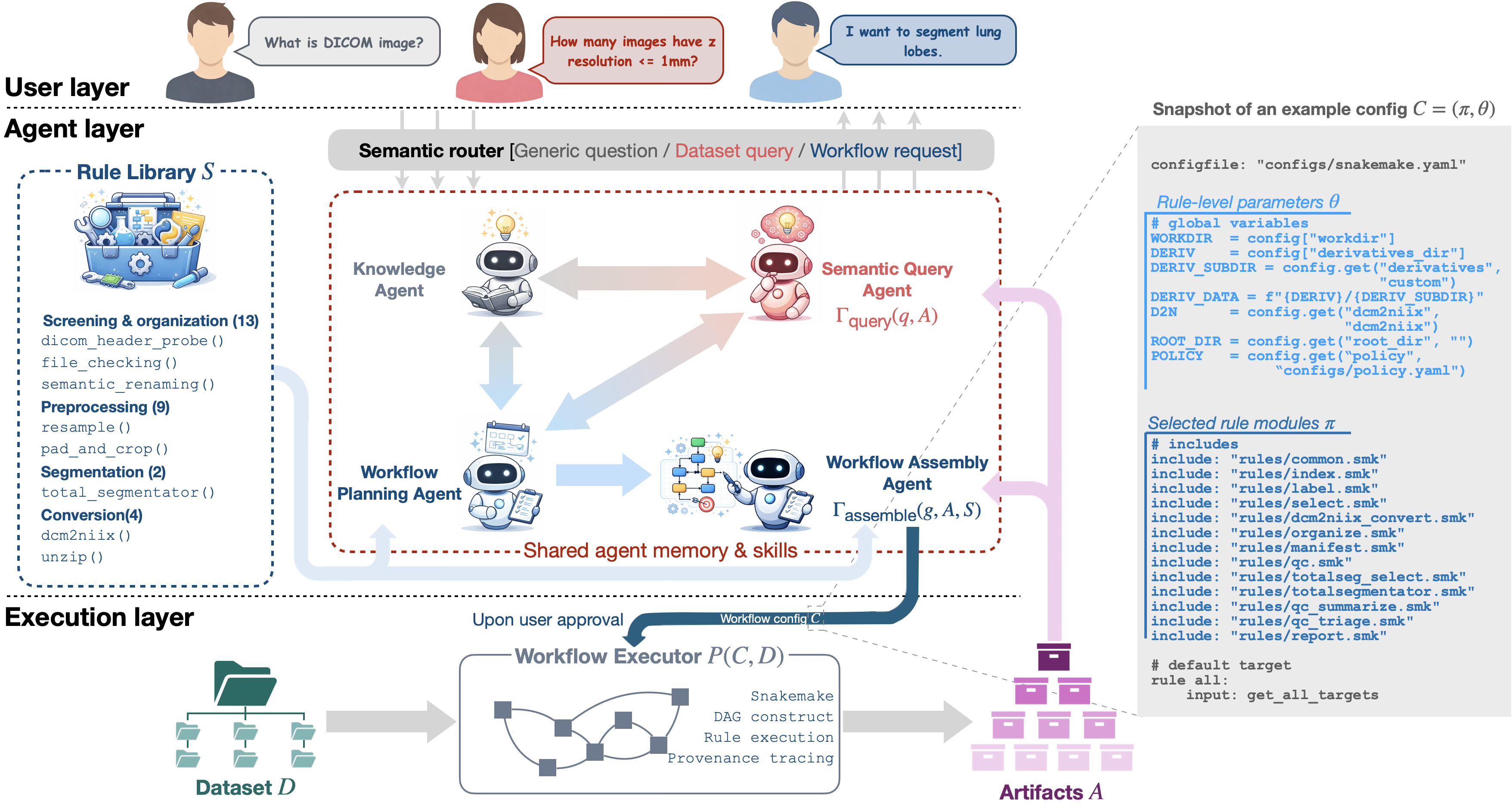}
    \caption{Overview of the proposed agentic system. User requests are routed to either (1) Knowledge Agent (generic questions), (2) a Semantic Query Agent operating on artifacts $\mathcal{A}$, or (3) a workflow pathway for analytical goals. The Workflow Planning and Assembly Agents construct a configuration $C$ from rule library $\mathcal{S}$ and artifacts $\mathcal{A}$, which is executed by a workflow manager to transform dataset $\mathcal{D}$ into artifacts $\mathcal{A}$.}
    \label{fig:method}
\end{figure}
We introduce a constrained artifact-based agent layer (Figure~\ref{fig:method}) that operates above deterministic workflow execution. 
Each artifact $a_j \in \mathcal{A}$ is defined under an artifact contract as a tuple
$a_j = (t_j, \phi_j, \psi_j)$,
where $t_j$ denotes artifact type, $\phi_j$ denotes structured attributes (e.g., modality, resolution, measurements), and $\psi_j$ encodes provenance information including generating rule and dependency relationships. 
This contract provides a formally defined, machine-readable state representation over which semantic reasoning is grounded.
On top of $\mathcal{A}$, the agent layer implements two constrained functions:
$\textbf{Workflow Assembly:} \quad {\Gamma}_\text{assemble} (g, \mathcal{A}, \mathcal{S}) = C$,
which synthesizes a configuration $C$ by selecting and composing rules from $\mathcal{S}$ conditioned on the current artifact state and analytical goal $g$;
$\textbf{Semantic Querying:} \quad {\Gamma}_\text{query} (q, {\mathcal{A}}) = r$,
which returns a response $r$ derived strictly from contract-compliant artifacts to address the user's query $q$.
These operators form a constrained agentic loop where planning and querying are performed over structured state, while execution remains delegated to the deterministic workflow executor $P(\mathcal{D}, C)$~(Figure~\ref{fig:method}).

\noindent\textbf{Adaptation via Workflow Assembly}
To deploy a workflow on a new dataset $\mathcal{D}_{\text{new}}$, the system first performs an inspection skill to identify available modalities, file organization, and prior processing status. 
This inspection summarize observable dataset properties and becomes part of the artifact state.

Given an initial analytical request $g_0$, the planning component evaluates whether the objective is well-defined with respect to the current artifact state and available rule library $\mathcal{S}$. 
Because users may not fully know dataset characteristics or available tools, the agent may iteratively refine the request into a structured and executable goal $g$ by clarifying required inputs, outputs, and processing assumptions. 
Once a feasible goal $g$ is established, $\Gamma_{\text{assemble}}$ selects rules from $\mathcal{S}$ whose declared inputs are satisfied by existing artifacts and whose outputs fulfill the objective. 
Dependency constraints are enforced during selection, and rule-level parameters are assigned to produce an explicit configuration
$C = (\pi, \theta)$, as shown in Figure~\ref{fig:method}.
The assembled workflow and directed acyclic graph~(DAG) are then presented to the user for approval before execution. 
Adaptation therefore occurs through goal planning and configuration synthesis, while computation remains deterministic.

\noindent \textbf{Semantic Query} 
The semantic query function enables inspection of workflow state without re-accessing raw files. 
Natural-language queries are interpreted as constraints over artifact attributes $\phi_j$ and provenance $\psi_j$. 
For example, a query about available T1-weighted scans with specific resolution is translated into filters over modality and voxel spacing fields stored in artifacts.
All responses are generated exclusively from contract-compliant artifacts and reference explicit $\phi_j$ and $\psi_j$. 
Because the language model operates only over structured artifact fields, responses are grounded in recorded workflow state rather than inferred from unstructured data.

\noindent \textbf{Reproducibility through Workflow Executor} 
Once a configuration $C$ is generated, execution is delegated to a workflow executor~(\texttt{Snakemake}~\cite{koster2012snakemake} in our implementation). 
The executor constructs a DAG from declared rule inputs and outputs, ensuring that dependencies are resolved prior to execution. 
Given fixed dataset $\mathcal{D}$, configuration $C$, and computational environment, execution is deterministic.
Intermediate and final artifacts are produced only when declared inputs are satisfied, and rule-level provenance is recorded automatically. 

\noindent \textbf{Implementation Details}
The agent operates locally via \texttt{Ollama} using the DeepSeek-R1 (14B) model in inference mode. 
Workflow rules are organized as modular \texttt{.smk} components and registered in a rule catalog that explicitly declares inputs, outputs, and parameter interfaces. 
Artifacts are stored as structured \texttt{.json} or \texttt{.csv} records consistent with the artifact contract. 
Workflow execution is performed using Snakemake (v7.x), which constructs the DAG from the generated configuration and executes tasks with file-based dependency tracking. 
Experiments were conducted on a Linux workstation equipped with an NVIDIA A6000 GPU for downstream model inference. 

\section{Experiments and Results}
\begin{table*}[t]
\centering
\caption{Dataset characteristics and quantitative evaluation of  reproducibility and adaptability. For each dataset and analytical goal, we report the number of rules in the domain-expert ground-truth configuration, reproducibility via DAG equivalence across repeated executions, and adaptability metrics$^\dagger$. }
\resizebox{\textwidth}{!}{%
\begin{tabular}{cccccccccccc}
\toprule
\multicolumn{6}{c}{\bf Dataset Characterization} & 
\multicolumn{2}{c}{\bf Assembly} & 
\multicolumn{1}{c}{\bf Reprod.} & 
\multicolumn{3}{c}{\bf Adaptability} \\
\cmidrule(lr){1-6}
\cmidrule(lr){7-8}
\cmidrule(lr){9-9}
\cmidrule(lr){10-12}
\scriptsize{Dataset} & \scriptsize{Modality} & \scriptsize{Cohort} & \scriptsize{Heteroge.} & \scriptsize{\#subj} & \scriptsize{\#ses} & 
\scriptsize{Analytical goal} & \scriptsize{\#Rules} & \scriptsize{DAG} & 
\scriptsize{IRM(\%)$\uparrow$} & \scriptsize{PL}$\downarrow$ & \scriptsize{FO(\%)$\uparrow$}  \\
\midrule

\multirow{3}{*}{\texttt{NLST}~\cite{kramer2011lung}} & \multirow{3}{*}{CT} & \multirow{3}{*}{Research} &
\multirow{3}{*}{Mid.} & \multirow{3}{*}{$200$} & \multirow{3}{*}{$592$}
& convert+curate & 11 & $\checkmark$ & $100$ & 1 & $98.8$ \\

& & & & & &
\cellcolor{lightgray}lung lobe seg &
\cellcolor{lightgray}14 &
\cellcolor{lightgray}$\checkmark$ &
\cellcolor{lightgray}$85.7$ &
\cellcolor{lightgray}4 &
\cellcolor{lightgray}$97.9$  \\
& & & & & &
lung nodule seg & 14 & $\checkmark$ & $100$ & 3 & $97.4$ \\
\midrule

\multirow{3}{*}{\texttt{LungCaTrial}} & \multirow{3}{*}{CT} & \multirow{3}{*}{Clinical} &
\multirow{3}{*}{High} & \multirow{3}{*}{636} & \multirow{3}{*}{$2,950$} & 
convert+curate & 12 & $\checkmark$ & $100$ & 2 &  $99.0$ \\

& & & & & &
\cellcolor{lightgray}lung nodule seg &
\cellcolor{lightgray}15 &
\cellcolor{lightgray}$\checkmark$ & 
\cellcolor{lightgray}$100$ &
\cellcolor{lightgray}4 &
\cellcolor{lightgray}$90.7$\\

& & & & & &
lung cancer risk & 17 & $\checkmark$ & $88.2$ & 6 &  $91.1$ \\
\midrule

\multirow{3}{*}{\texttt{BrainICU}} & \multirow{3}{*}{MRI} & \multirow{3}{*}{Research} &
\multirow{3}{*}{Mid.} & \multirow{3}{*}{$113$} & \multirow{3}{*}{$135$} &
convert+curate & 17 & $\checkmark$ & $94.1$ & 5 &  $100$ \\

& & & & & &
\cellcolor{lightgray}harmonization &
\cellcolor{lightgray}19 &
\cellcolor{lightgray}$\checkmark$ &
\cellcolor{lightgray}$100$&
\cellcolor{lightgray}8 &
\cellcolor{lightgray}$97.7$ \\

& & & & & & 
brain seg & 20 & $\checkmark$ & $100$ & 8 &  $96.3$ \\
\midrule
\midrule
\multicolumn{12}{p{1.3\textwidth}}{
 \textbf{Example: Goal Planning} (\texttt{NLST}, lung nodule seg)\newline
{\color{blue}\texttt{User > I want to segment the lung nodules in this dataset.}} \newline
\texttt{Agent> Current status: [omit to save space]. Suggested rules: [omit to save space]. Assumptions made: [omit to save space]. Needs confirmation: 1. Kernel. I see multiple kernels of the same session. Do you want to segment them all? 2. [omit to save space] ...} \newline
{\color{blue}\texttt{User > 1. Segment them all. 2. [omit to save space] ...\newline }}
$\cdots$ \newline
\texttt{Agent> This is an updated workflow with rules [omit to save space]. Assumptions made: [omit to save space]. If you approve, I will assemble and execute.}
}
\\
\bottomrule
\end{tabular}%
}
\vspace{2mm}
\begin{minipage}{\textwidth}
\tiny
$\dagger$\textbf{Abbreviations:}
\textbf{IRM:} Initial Rule Matching. Percentage of rules matched between agent's first selection and ground truth;
\textbf{PL:} Number of planning iterations (chats from user) before a workflow is assembled by the agent; 
\textbf{FO:} Percentage of sessions whose final outputs satisfy the intended goal against manually curated results.
\end{minipage}
\label{tab:reprod_adapt}
\end{table*}
We evaluate whether the proposed framework (i) preserves deterministic execution under repeated runs, (ii) assembles goal-conditioned workflows across varying cohorts, and (iii) supports artifact-grounded semantic query without re-running. 
Experiments were conducted on three cohorts including both research and clinical settings (Table~\ref{tab:reprod_adapt}): \texttt{NLST} (research CT; moderate heterogeneity), \texttt{LungCaTrial} (clinical trial CT; high heterogeneity), and \texttt{BrainICU} (research MRI; moderate heterogeneity). 
Across cohorts, we test our method under multiple analytical goals including data conversion/curation and downstream analysis (i.e., lung lobe and nodule segmentation~\cite{wasserthal2023totalsegmentator}, cancer risk estimation~\cite{li2023time}, harmonization~\cite{zuo2023haca3}, and brain segmentation~\cite{yu2023unest}), resulting in $9$ different  workflows.

\noindent\textbf{Reproducibility}
For each dataset and analytical goal, we invoke the agent twice after clearing memory and intermediate states, then compare the resulting workflow graphs. 
Reproducibility is measured by DAG equivalence between the two runs. 
Across all settings, the induced DAGs were identical (Table~\ref{tab:reprod_adapt}), indicating that, under fixed goals and environment, the planning-to-configuration pathway yields consistent executable workflows and deterministic execution graphs.

\noindent \textbf{Adaptability}
We measure how well the agent adapts workflows to each cohort and analytical goal using three metrics. 
First, \underline{Initial Rule Matching (IRM)} quantifies overlap between the agent’s initial proposed rule set and a domain-expert configuration, reflecting the quality of dataset-aware planning immediately after inspection. 
Second, \underline{Planning Iterations (PL)} counts the number of user-agent exchanges before finalizing a configuration, capturing interaction cost. 
Third, \underline{Final Output (FO)} reports the percentage of sessions whose outputs satisfy the intended goal against manually curated results.

As shown in Table~\ref{tab:reprod_adapt}, the agent’s initial plans are close to expert-defined pipelines, with  IRM$>85\%$ across all cohorts and analytical goals.
Additionally, iterative planning resolves intent ambiguity (e.g., which series to target), as shown in Table~\ref{tab:reprod_adapt} example.
This is reflected by modest PL values (a workflow is accurately assembled after $1$ to $8$ rounds of chats between agent and user) even in the clinical cohorts, while maintaining high session-level success in most tasks. 
We then evaluated all failure cases: mismatches are concentrated in cases of missing/corrupted data and ambiguous series selection (e.g., multiple candidate series satisfying similar constraints), which typically require manual intervention even in conventional workflows.

\noindent \textbf{Semantic Query}
\begin{table*}[t]
\centering
\caption{Semantic query evaluation (\texttt{deepseek-r1:14b} and \texttt{qwen2.5:14b}) and ablation experiment with/without artifact contract. Accuracy denotes exact-match percentage over 20 queries per category.}
\label{tab:semantic_query}

\begin{minipage}[t]{0.50\textwidth}
\centering
\vspace{0pt}
\resizebox{\textwidth}{!}{
\begin{tabular}{cccccc}
\toprule
\textbf{Dataset} & \textbf{Model} & \textbf{Contract?} & \textbf{Status} & \textbf{Filter} & \textbf{Provenance} \\
\midrule

\multirow{3}{*}{\texttt{NLST}} 
& \texttt{DeepSeek} & \xmark & 95\% & 20\% & 80\% \\
& \cellcolor{lightgray}\texttt{Qwen} 
  & \cellcolor{lightgray}$\checkmark$ 
  & \cellcolor{lightgray}100\%
  & \cellcolor{lightgray}90\%
  & \cellcolor{lightgray}100\% \\
& \texttt{DeepSeek} & $\checkmark$ & {\color{blue}100\%} & {\color{blue}95\%} & {\color{blue}100\%} \\
\midrule

\multirow{3}{*}{\texttt{LungCaTrial}} 
& \texttt{DeepSeek} & \xmark & 90\% & 15\% & 80\% \\
& \cellcolor{lightgray}\texttt{Qwen} 
  & \cellcolor{lightgray}$\checkmark$
  & \cellcolor{lightgray}100\%
  & \cellcolor{lightgray}85\%
  & \cellcolor{lightgray}100\% \\
& \texttt{DeepSeek} & $\checkmark$ & {\color{blue}100\%} & {\color{blue}90\%} & {\color{blue}100\%} \\
\midrule

\multirow{3}{*}{\texttt{BrainICU}} 
& \texttt{DeepSeek} & \xmark & 95\% & 10\% & 75\% \\
& \cellcolor{lightgray}\texttt{Qwen} 
  & \cellcolor{lightgray}$\checkmark$
  & \cellcolor{lightgray}95\%
  & \cellcolor{lightgray}90\%
  & \cellcolor{lightgray}100\% \\
& \texttt{DeepSeek} & $\checkmark$ & {\color{blue}100\%} & {\color{blue}90\%} & {\color{blue}100\%} \\
\bottomrule
\end{tabular}
}
\end{minipage}
\hfill
\begin{minipage}[t]{0.45\textwidth}
\vspace{0pt}
\scriptsize
\setlength{\fboxsep}{6pt}
\setlength{\fboxrule}{0.4pt}
\fbox{%
\begin{minipage}[t]{0.96\textwidth}
\raggedright
\textbf{Example (Filter/counting)}\\[1mm]
{\color{blue}\texttt{User > How many NIfTI images were acquired using a Siemens scanner and have slice thickness greater than 1\,mm?}}\\[1mm]
\texttt{Agent> 23. File paths can be retrieved from artifact records in \texttt{data\_inventory.csv} and \texttt{qa\_report.csv}}.
\end{minipage}}
\end{minipage}

\end{table*}
We evaluate semantic querying across three categories: status, filtering/counting, and provenance. 
For each category, $20$ representative questions are evaluated using exact-match accuracy against ground truth.
We used two local LLMs in this experiment~(Table~\ref{tab:semantic_query}).
Status queries are generally answered correctly, as they simply require checking whether certain files exist (e.g., “Has lung nodule segmentation been executed for subject X?”). 
Filtering/counting queries are more demanding because they require structured attribute reasoning, such as checking specific voxel resolution or scanner manufacturer. 
Provenance queries require tracing rule-level dependencies, for example identifying which preprocessing steps generated a particular result.
The proposed method accurately answers all questions in status and provenance questions, and only made one to two mistakes in dataset query questions.
Upon investigating error cases, we found that most failures were caused by incomplete or corrupted DICOM headers, leading to missing or inaccurate artifact attributes $\phi_j$. 
For example, body part was inferred from DICOM headers, and this process can be unreliable. 
These errors therefore reflect limitations of upstream metadata rather than ungrounded generation. 
Incorporating additional skills, such as image-based body-part regression, may further improve robustness.

\noindent \textbf{Ablation Study} 
To isolate the effect of artifact contract, we perform an ablation in which the agent is denied access to the artifact registry and instead receives only directory listings and raw filenames, mimicking a conventional manual scripting workflow. 
The same query set is evaluated under this condition.

After removing the artifact contract, the largest degradation occurs in filtering/counting queries. 
For example, when asked to identify sessions with voxel resolution below a specified threshold, filename-based reasoning fails because resolution is not encoded in directory names; it is only available in structured artifact fields extracted during preprocessing.
Provenance queries also degrade without artifact access. When asked which rule generated a particular segmentation output, the grounded system retrieves the recorded rule identifier and dependency chain from $\psi_j$, whereas the filename-only condition lacks explicit traceability and must rely on heuristic inference.
These results demonstrate the importance of contract-compliant artifacts.
The artifact contract therefore serves not only as a reproducibility mechanism but also as a grounding interface that reduces inference over unstructured file organization.

\section{Discussion and Conclusion}
We presented an artifact contract-based agent framework for adaptive and reproducible medical image processing across heterogeneous research and clinical datasets. 
The system separates semantic planning from deterministic execution: workflow configurations are synthesized at the semantic layer, while execution remains reproducible through a workflow manager with explicit provenance tracking.
Experiments show that configurations can be generated across diverse CT and MRI cohorts, and that repeated runs produce identical DAGs. 
In our ablation study, we show that the artifact contract improves the semantic query accuracy.
The main limitation arises from incomplete artifact attributes. 
In our experiments, most query errors were caused by missing or corrupted metadata, leading to inaccurate artifact fields. 
This highlights the dependence of semantic query on upstream data quality. 
Similarly, adaptability is bounded by the coverage of the rule catalog; unsupported analytical objectives require additional rules.

Nevertheless, the framework is designed for extension.
New rules and skills can be added without altering the overall architecture, enabling progressive enrichment of the artifact schema and planning capabilities.
By grounding interaction in structured workflow state and preserving deterministic execution, the proposed system provides a practical and extensible foundation for adaptive medical image analysis in real-world settings.

\section*{Acknowledgment}
We sincerely thank Adam~Saunders and Michael~E.~Kim for sharing their experience with \texttt{Snakemake}, which contributed to the development of this work.

This research was funded by the National Cancer Institute (NCI) grant R01 CA253923‐04 and R01 CA 253923‐04S1. This work was also supported by the following awards: National Science Foundation CAREER 1452485; NCI U01 CA196405; UL1 RR024975‐01 from the National Center for Research Resources and UL1 TR000445‐06 from the National Center for Advancing Translational Sciences; the Martineau Innovation Fund grant through the Vanderbilt‐Ingram Cancer Center Thoracic Working Group; and NCI Early Detection Research Network grant 2U01CA152662. 
Additional support was provided by U01CA152662, SNF300329~(Helmsley), and P50HD103537.

During the development of this work, we used generative AI to create code segments from task descriptions and to assist with debugging, editing, and code autocompletion. In addition, generative AI technologies were used to help structure sentences and perform grammatical checks. We emphasize that the conceptualization, ideation, and all prompts provided to the AI originated entirely from the authors’ intellectual and creative contributions. The authors take full responsibility for reviewing and validating all AI-generated content included in this work.
%
%
%
\bibliographystyle{splncs04}
\bibliography{MICCAI2026-Latex-Template/own-bib}

@article{zuo2023haca3,
  title={{HACA3: A Unified Approach for Multi-site MR Image Harmonization}},
  author={Zuo, Lianrui and Liu, Yihao and Xue, Yuan and Dewey, Blake E and Remedios, Samuel W and Hays, Savannah P and Bilgel, Murat and Mowry, Ellen M and Newsome, Scott D and Calabresi, Peter A and others},
  journal={Computerized Medical Imaging and Graphics},
  volume={109},
  pages={102285},
  year={2023},
  publisher={Elsevier}
}

@article{chen2025beyond,
  title={{Beyond the LUMIR challenge: The pathway to foundational registration models}},
  author={Chen, Junyu and Wei, Shuwen and Honkamaa, Joel and Marttinen, Pekka and Zhang, Hang and Liu, Min and Zhou, Yichao and Tan, Zuopeng and Wang, Zhuoyuan and Wang, Yi and others},
  journal={arXiv preprint arXiv:2505.24160},
  year={2025}
}

@inproceedings{zhang2024harmonization,
  title={{Harmonization-Enriched Domain Adaptation with Light Fine-tuning for Multiple Sclerosis Lesion Segmentation}},
  author={Zhang, Jinwei and Zuo, Lianrui and Dewey, Blake E and Remedios, Samuel W and Hays, Savannah P and Pham, Dzung L and Prince, Jerry L and Carass, Aaron},
  booktitle={Medical Imaging 2024: Clinical and Biomedical Imaging},
  volume={12930},
  pages={633--639},
  year={2024},
  organization={SPIE}
}

@article{ferrucci2023baltimore,
  title={{Baltimore Longitudinal Study of Aging (BLSA)}},
  author={Ferrucci, Luigi and Resnick, Susan M and Deal, Jennifer A},
  journal={Hearing Loss Rehabilitation and Higher-Order Auditory and Cognitive Processing},
  pages={116},
  year={2023},
  publisher={Frontiers Media SA}
}

@article{lamontagne2019oasis,
  title={{OASIS-3: longitudinal neuroimaging, clinical, and cognitive dataset for normal aging and Alzheimer disease}},
  author={LaMontagne, Pamela J and Benzinger, Tammie LS and Morris, John C and Keefe, Sarah and Hornbeck, Russ and Xiong, Chengjie and Grant, Elizabeth and Hassenstab, Jason and Moulder, Krista and Vlassenko, Andrei G and others},
  journal={medrxiv},
  pages={2019--12},
  year={2019},
  publisher={Cold Spring Harbor Laboratory Press}
}

@article{naseer2022performance,
  title={{Performance analysis of state-of-the-art CNN architectures for LUNA16}},
  author={Naseer, Iftikhar and Akram, Sheeraz and Masood, Tehreem and Jaffar, Arfan and Khan, Muhammad Adnan and Mosavi, Amir},
  journal={Sensors},
  volume={22},
  number={12},
  pages={4426},
  year={2022},
  publisher={MDPI}
}

@article{menze2014multimodal,
  title={{The multimodal brain tumor image segmentation benchmark (BRATS)}},
  author={Menze, Bjoern H and Jakab, Andras and Bauer, Stefan and Kalpathy-Cramer, Jayashree and Farahani, Keyvan and Kirby, Justin and Burren, Yuliya and Porz, Nicole and Slotboom, Johannes and Wiest, Roland and others},
  journal={IEEE transactions on medical imaging},
  volume={34},
  number={10},
  pages={1993--2024},
  year={2014},
  publisher={IEEE}
}

@article{poldrack2024past,
  title={{The past, present, and future of the brain imaging data structure (BIDS)}},
  author={Poldrack, Russell A and Markiewicz, Christopher J and Appelhoff, Stefan and Ashar, Yoni K and Auer, Tibor and Baillet, Sylvain and Bansal, Shashank and Beltrachini, Leandro and Benar, Christian G and Bertazzoli, Giacomo and others},
  journal={Imaging Neuroscience},
  volume={2},
  pages={imag--2},
  year={2024},
  publisher={MIT Press One Broadway, 12th Floor, Cambridge, Massachusetts 02142, USA~…}
}

@article{koster2012snakemake,
  title={Snakemake—a scalable bioinformatics workflow engine},
  author={K{\"o}ster, Johannes and Rahmann, Sven},
  journal={Bioinformatics},
  volume={28},
  number={19},
  pages={2520--2522},
  year={2012},
  publisher={Oxford University Press}
}

@misc{kramer2011lung,
  title={{Lung cancer screening with low-dose helical CT: results from the National Lung Screening Trial (NLST)}},
  author={Kramer, Barnett S and Berg, Christine D and Aberle, Denise R and Prorok, Philip C},
  journal={Journal of medical screening},
  volume={18},
  number={3},
  pages={109--111},
  year={2011},
  publisher={SAGE Publications Sage UK: London, England}
}

@article{peng2011reproducible,
  title={Reproducible research in computational science},
  author={Peng, Roger D},
  journal={Science},
  volume={334},
  number={6060},
  pages={1226--1227},
  year={2011},
  publisher={American Association for the Advancement of Science}
}

@article{stodden2016enhancing,
  title={Enhancing reproducibility for computational methods},
  author={Stodden, Victoria and McNutt, Marcia and Bailey, David H and Deelman, Ewa and Gil, Yolanda and Hanson, Brooks and Heroux, Michael A and Ioannidis, John PA and Taufer, Michela},
  journal={Science},
  volume={354},
  number={6317},
  pages={1240--1241},
  year={2016},
  publisher={American Association for the Advancement of Science}
}

@article{di2017nextflow,
  title={Nextflow enables reproducible computational workflows},
  author={Di Tommaso, Paolo and Chatzou, Maria and Floden, Evan W and Barja, Pablo Prieto and Palumbo, Emilio and Notredame, Cedric},
  journal={Nature biotechnology},
  volume={35},
  number={4},
  pages={316--319},
  year={2017},
  publisher={Nature Publishing Group US New York}
}

@inproceedings{yao2022react,
  title={React: Synergizing reasoning and acting in language models},
  author={Yao, Shunyu and Zhao, Jeffrey and Yu, Dian and Du, Nan and Shafran, Izhak and Narasimhan, Karthik R and Cao, Yuan},
  booktitle={The eleventh international conference on learning representations},
  year={2022}
}

@article{schick2023toolformer,
  title={Toolformer: Language models can teach themselves to use tools},
  author={Schick, Timo and Dwivedi-Yu, Jane and Dess{\`\i}, Roberto and Raileanu, Roberta and Lomeli, Maria and Hambro, Eric and Zettlemoyer, Luke and Cancedda, Nicola and Scialom, Thomas},
  journal={Advances in neural information processing systems},
  volume={36},
  pages={68539--68551},
  year={2023}
}

@inproceedings{park2023generative,
  title={Generative agents: Interactive simulacra of human behavior},
  author={Park, Joon Sung and O'Brien, Joseph and Cai, Carrie Jun and Morris, Meredith Ringel and Liang, Percy and Bernstein, Michael S},
  booktitle={Proceedings of the 36th annual acm symposium on user interface software and technology},
  pages={1--22},
  year={2023}
}

@article{wasserthal2023totalsegmentator,
  title={{TotalSegmentator: robust segmentation of 104 anatomic structures in CT images}},
  author={Wasserthal, Jakob and Breit, Hanns-Christian and Meyer, Manfred T and Pradella, Maurice and Hinck, Daniel and Sauter, Alexander W and Heye, Tobias and Boll, Daniel T and Cyriac, Joshy and Yang, Shan and others},
  journal={Radiology: Artificial Intelligence},
  volume={5},
  number={5},
  pages={e230024},
  year={2023},
  publisher={Radiological Society of North America}
}

@inproceedings{li2023time,
  title={Time-distance vision transformers in lung cancer diagnosis from longitudinal computed tomography},
  author={Li, Thomas Z and Xu, Kaiwen and Gao, Riqiang and Tang, Yucheng and Lasko, Thomas A and Maldonado, Fabien and Sandler, Kim L and Landman, Bennett A},
  booktitle={Medical Imaging 2023: Image Processing},
  volume={12464},
  pages={229--238},
  year={2023},
  organization={SPIE}
}

@article{yu2023unest,
  title={Unest: local spatial representation learning with hierarchical transformer for efficient medical segmentation},
  author={Yu, Xin and Yang, Qi and Zhou, Yinchi and Cai, Leon Y and Gao, Riqiang and Lee, Ho Hin and Li, Thomas and Bao, Shunxing and Xu, Zhoubing and Lasko, Thomas A and others},
  journal={Medical Image Analysis},
  volume={90},
  pages={102939},
  year={2023},
  publisher={Elsevier}
}
%




\end{document}